\documentclass[conference]{IEEEtran}
\usepackage[utf8]{inputenc}
\usepackage[T1]{fontenc}
\usepackage{hyperref}
\usepackage{url}
\usepackage{graphicx}
\usepackage{amsmath}
\usepackage{booktabs}
\usepackage{tabularx}
\usepackage{morefloats}
\usepackage{listings}
\usepackage{xcolor}
\usepackage{textcomp}

\hypersetup{
    colorlinks=true,
    linkcolor=blue,
    filecolor=magenta,
    urlcolor=cyan,
    citecolor=blue
}

\begin{document}

\title{MCP vs CLI for AI Agents: A Comparative Analysis of Tool Integration Strategies for Autonomous Systems}

\author{\IEEEauthorblockN{Arun Malik, Nitish Talasu, Bradley Klemick, Vineet Tushar Trivedi,\\
Prachi Shah, Sanay Devi, Junjie Zhang, Kunal Sonawane}
\IEEEauthorblockA{Microsoft Corporation, Redmond, WA, USA\\
Email: arunma@microsoft.com}}

\maketitle

\begin{abstract}

        AI agents require interfaces to interact with external systems such as files, APIs, databases, and cloud services. Two competing paradigms have emerged: CLI (Command Line Interface), where agents execute shell commands directly, and MCP (Model Context Protocol), a structured JSON-RPC protocol providing governed, typed tool access. This paper presents a comparative analysis across 14 evaluation dimensions, supported by worked examples, security assessment against established compliance frameworks, and economic modeling. Our analysis indicates that CLI provides superior efficiency for single-user, local-only workflows, while MCP addresses requirements that CLI fundamentally cannot satisfy, including fine-grained access control, credential isolation, structured audit, and multi-tenant governance. We further examine how progressive disclosure techniques eliminate MCP's historical weakness (context window overhead), yielding token efficiency comparable to CLI while preserving its architectural benefits. A hybrid architecture is recommended for production systems. This analysis draws upon academic, government, and industry sources including NSA formal guidance (CSI U/OO/6030316-26), OWASP security standards, and economic modeling from BCG and Gartner.


\end{abstract}

\begin{table*}[t]
\centering
\small
\begin{tabularx}{\textwidth}{|X|X|X|}
\hline
\textbf{Area} & \textbf{Key Finding} & \textbf{Section} \\
\hline
Operational performance & CLI is faster for simple, local tasks (2--5$\times$ latency advantage); MCP p... & \S{}III,\S{}IV \\
Security and compliance & CLI grants unbounded system access; MCP satisfies 6 of 6 evaluated compliance... & \S{}V \\
Token economics & Naive MCP loading imposes significant overhead (up to 55K tokens); progressiv... & \S{}VII,\S{}VIII \\
Enterprise scalability & CLI requires container-per-user isolation at scale; MCP's session-level RBAC ... & \S{}III,\S{}VII \\
Risk analysis & CLI exposes 7 architectural attack vectors (injection, exfiltration, escalati... & \S{}VI \\
\hline
\end{tabularx}
\end{table*}

\begin{figure}[htbp]
\centering
\includegraphics[width=\columnwidth]{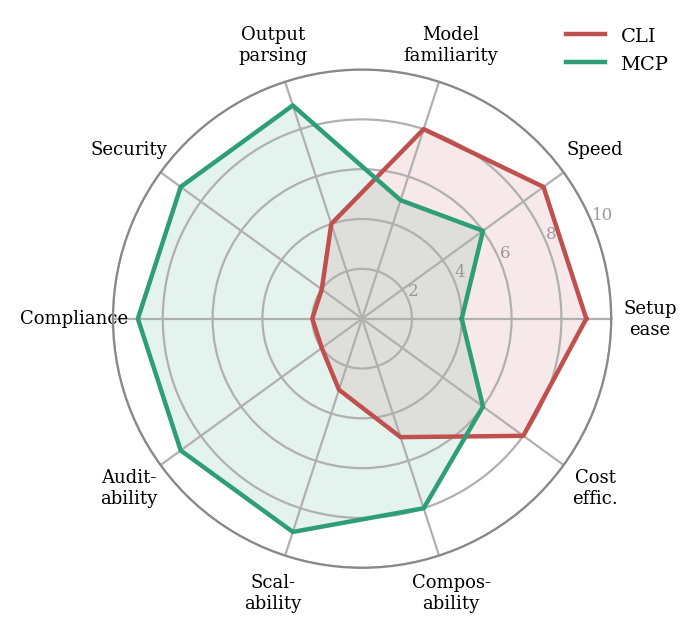}
\caption{Multi-dimensional comparison of CLI and MCP across ten evaluation dimensions (normalized 0--10; higher is better).}
\end{figure}

\section{Introduction}

    AI agents (LLM-based systems that autonomously decide which tools to call and in what sequence) require integration mechanisms to interact with external systems. As organizations deploy AI agents at enterprise scale, the choice of integration architecture has profound implications for security, compliance, cost, and operational reliability.

    Two paradigms compete for this role: (1) \textbf{CLI (Command Line Interface)}, where the agent executes shell commands directly on the host system, leveraging the model's pre-trained knowledge of Unix/Windows commands; and (2) \textbf{MCP (Model Context Protocol)}, a standardized JSON-RPC protocol connecting AI models to purpose-built tool servers with typed inputs/outputs, permission boundaries, and structured audit trails.

    This paper provides a rigorous comparative analysis answering the core research question: \textit{Is MCP a necessary abstraction layer for secure, governed AI-agent interactions at enterprise scale, or is it unnecessary complexity when CLI tools can accomplish the same tasks more efficiently?}

    Our principal conclusion is that the CLI-vs-MCP debate presents a false dichotomy. The two paradigms serve fundamentally different trust levels. The optimal architecture for production systems is hybrid: CLI for local developer operations, MCP for any interaction that crosses a trust boundary or requires organizational governance.

    The remainder of this paper is organized as follows. Section II defines key terms. Section III establishes scope and assumptions. Section IV presents the core 14-dimension comparative analysis. Section V provides worked examples at increasing complexity. Section VI evaluates security and compliance. Section VII examines risk. Section VIII models cost and token economics. Section IX addresses progressive disclosure. Section X surveys industry adoption. Section XI presents a decision framework, and Section XII concludes with recommendations. Table I summarizes the coverage structure.

\section{Preliminaries}
\begin{table*}[t]
\centering
\small
\begin{tabularx}{\textwidth}{|X|X|X|}
\hline
\textbf{Term} & \textbf{Definition} & \textbf{Key Characteristic} \\
\hline
CLI & Command Line Interface - text commands in bash/PowerShell/zsh & Unstructured I/O; full system access; model knowledge from training \\
MCP & Model Context Protocol - JSON-RPC connecting models to tool servers & Structured typed I/O; per-tool permissions; schema-defined contracts \\
AI Agent & LLM system autonomously selecting and invoking tools & Decisions based on context window contents; tool-use loop \\
Context Window & Total token budget for a conversation/task (128K--1M tokens) & Finite; tool schemas consume tokens; key economic constraint \\
Tool Schema & JSON definition of tool name, params, and return type & MCP requires at runtime; CLI relies on pre-trained knowledge \\
\hline
\end{tabularx}
\end{table*}

\section{Assumptions \& Scope}
\subsection{Assumptions}
\begin{enumerate}
  \item The AI model is a modern LLM (GPT-4+, Claude 3+, Gemini) with strong CLI knowledge from training data
  \item CLI access means unrestricted bash/shell execution on the host system
  \item MCP access means a properly configured MCP server with appropriate tool registrations
  \item \textbf{Cost encompasses tokens, latency, operational overhead, maintenance burden, and incident liability}, not just API charges
  \item Security requirements scale with organizational maturity and regulatory obligations
  \item Reliability includes error recovery, retry semantics, idempotency, and state management
\end{enumerate}

\subsection{Scope}
\begin{table*}[t]
\centering
\small
\begin{tabularx}{\textwidth}{|X|X|}
\hline
\textbf{In Scope} & \textbf{Out of Scope} \\
\hline
File operations, Git, API calls, database queries & GUI automation, browser-based tools \\
Single-agent and multi-agent patterns & Agent-to-agent communication protocols \\
Enterprise security, RBAC, compliance & Specific vendor pricing (changes frequently) \\
Token economics, latency, reliability, state management & Model fine-tuning approaches \\
\hline
\end{tabularx}
\end{table*}

\subsection{Evaluation Criteria}
Each paradigm is assessed against six dimensions:

\begin{table*}[t]
\centering
\small
\begin{tabularx}{\textwidth}{|X|X|}
\hline
\textbf{Criterion} & \textbf{Measures} \\
\hline
Efficiency & Token cost, latency, steps to complete, context window utilization \\
Security & Access control, blast radius, credential isolation, input validation \\
Reliability & Error rates, output determinism, retry semantics, idempotency \\
Scalability & Multi-user, multi-tenant, enterprise governance \\
Maintainability & Version brittleness, state leakage, upgrade path, schema evolution \\
Compliance & SOC2, HIPAA, GDPR, PCI-DSS, ISO 27001, FedRAMP \\
\hline
\end{tabularx}
\end{table*}

\section{Comparative Analysis: 14 Dimensions}
\begin{table*}[t]
\centering
\small
\begin{tabularx}{\textwidth}{|X|X|X|X|X|}
\hline
\textbf{\#} & \textbf{Dimension} & \textbf{CLI} & \textbf{MCP} & \textbf{Winner} \\
\hline
1 & Setup Cost & Zero - commands pre-exist & Server deployment + schema registration & CLI \\
2 & Context Window Cost & \textasciitilde{}0 tokens (model knows commands) & 100--55,000 tokens (static);\textasciitilde{}170 tokenswith progressive discl... & TIE* \\
3 & Model Familiarity & Millions of examples in training data & Must read schema at runtime & CLI \\
4 & Output Parsing & Unstructured text - model interprets & Structured JSON - deterministic & MCP \\
5 & Error Handling & Exit codes + stderr (ambiguous) & Typed error responses with codes & MCP \\
6 & Access Control (RBAC) & All-or-nothing shell access & Per-tool, per-user, per-resource & MCP\S{}VI \\
7 & Audit Trail & Shell history (unreliable, unstructured) & Structured: who, what, when, result & MCP\S{}VI \\
8 & Blast Radius & Unlimited (rm -rf, credential theft) & Bounded to exposed capabilities & MCP\S{}VI \\
9 & Compliance & Very difficult - no built-in controls & Native: consent, classification, retention & MCP\S{}VI \\
10 & Multi-Tenant & Container per user (\$\$\$) & Session isolation, per-user scoping & MCP\S{}V-C \\
11 & Tool Discovery & No runtime discovery & Dynamic capability negotiation & MCP\S{}IX \\
12 & Composability & Pipe chains (fragile parsing) & Typed chaining with validation & MCP \\
13 & Speed (Simple Tasks) & Fastest - direct execution & JSON-RPC overhead (\textasciitilde{}50--200ms) & CLI \\
14 & Ecosystem Breadth & Thousands of Unix/Windows commands & 5,800+ MCP connectors (growing) & TIE \\
\hline
\end{tabularx}
\end{table*}

\begin{figure}[htbp]
\centering
\includegraphics[width=\columnwidth]{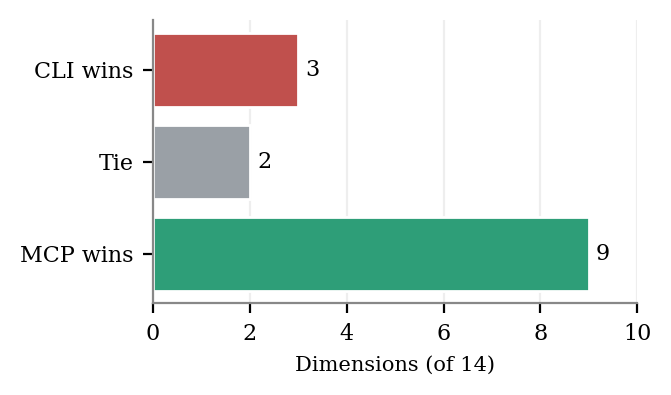}
\caption{Dimension wins by paradigm across 14 evaluated dimensions: CLI leads on setup, speed, and cost; MCP leads on security, governance, and scale.}
\end{figure}

\subsection{Dimension Analysis}
\textbf{Dimension 1: Setup Cost.} CLI commands exist on every Unix and Windows installation. An agent can invoke \texttt{grep}, \texttt{curl}, or \texttt{git} with no prior configuration. MCP requires deploying at least one server process (or connecting to a hosted endpoint), registering tool schemas, and configuring authentication. For ephemeral tasks on a developer's local machine, this overhead is difficult to justify.

\textbf{Dimension 3: Model Familiarity.} Large language models are trained on corpora containing millions of shell command examples from man pages, Stack Overflow posts, and open-source repositories. The model already knows that \texttt{grep -rn "pattern" .} searches recursively with line numbers. MCP tools, by contrast, are novel to the model; it must parse the JSON schema at runtime to learn what parameters a tool accepts. This distinction narrows over time as MCP schemas appear in training data, but as of mid-2026, CLI retains a significant advantage in zero-shot accuracy for shell operations.

\textbf{Dimension 4: Output Parsing.} CLI output is free-form text. The output of \texttt{ls -la} differs between GNU coreutils and BSD; \texttt{docker ps} output changed format between Docker 23 and 24. The model must interpret column alignment, handle locale-dependent date formats, and distinguish informational output from errors. MCP returns typed JSON with a contract: a \texttt{read\_file} tool always returns \texttt{\{"content": "..."\}}. Downstream processing (chaining tools, populating UI) becomes deterministic rather than heuristic.

\textbf{Dimension 12: Composability.} CLI achieves composition through pipes: \texttt{cat file | grep pattern | wc -l}. This works well for text streams but breaks when output format changes between versions. If an upstream command adds a header line, downstream counts become incorrect. MCP supports typed chaining: one tool's structured output feeds directly into another tool's validated input schema. The MCP client can verify type compatibility before invocation, preventing silent failures.

\textbf{Dimension 13: Speed.} For simple local operations, CLI executes a system call directly. No serialization, no network round-trip, no protocol negotiation. MCP adds JSON-RPC serialization, transport (typically stdio or HTTP), server-side deserialization, execution, and response serialization. Measured overhead ranges from 50ms (stdio transport, local server) to 200ms (HTTP transport, remote server). For a single file read, this latency is noticeable. For a complex workflow with 20+ tool calls, it becomes negligible relative to LLM inference time.

\textbf{Dimension 14: Ecosystem Breadth.} CLI has decades of accumulated tooling: thousands of Unix utilities, package managers, cloud CLIs (aws, az, gcloud), database clients, and container runtimes. MCP, launched in late 2024, has grown to over 5,800 registered connectors as of mid-2026. The ecosystems overlap substantially (most CLI tools now have MCP equivalents), making this dimension a draw. The relevant difference is not breadth but access model: CLI tools grant full capability by default; MCP connectors expose only explicitly declared operations.

\section{Worked Examples}
\subsection{File Operations (Trivial Task)}
\begin{quote}
\small
CLIcatnotes.mdgrep-rn"agent"*.mdSchema tokens:0Result:Raw textMCP\{"tool":"read\_file","params": \{"path":"notes.md"\} \}
\{"tool":"search","params": \{"query":"agent"\} \}Schema tokens:\textasciitilde{}2,000Result:Structured JSON
\end{quote}
\textit{Finding 1:} For trivial local operations, CLI is simpler and equally effective. The absence of protocol overhead results in lower latency and zero schema cost.

\subsection{Production Database Query}
\begin{quote}
\small
CLIpsql-hprod-db-Uadmin \textbackslash\{\}-c"SELECT * FROM customers WHERE region='EU'"Risks:DBA credentials exposed, DROP/DELETE possible, no rate limiting, connection strings in historyMCP\{"tool":"query\_database","params": \{"sql":"SELECT...","readonly": true\} \}Protections:Read-only enforced, DDL blocked, credentials opaque, row-level security, rate limited
\end{quote}
\textit{Finding 2:} MCP is a prerequisite for production data access. Any CLI-based approach exposes credentials and permits unrestricted DDL operations, which would fail a security audit under SOC2 or HIPAA requirements.

\subsection{Multi-User Enterprise (50 developers)}
\begin{quote}
\small
CLISeparate container/VM per user (\$\$\$)OS-level groups onlyAudit = grep bash\_historyCompliance: manual spreadsheetsMCPSingle server + OAuth/SSOPer-user RBAC matrixStructured logs $\rightarrow$ SIEMAutomated SOC2/HIPAA reports
\end{quote}
\textit{Finding 3:} At enterprise scale (multi-user, governed environments), MCP provides native governance primitives while CLI requires expensive infrastructure workarounds that still cannot match MCP's granularity.

\section{Security, Compliance \& Governance}
\subsection{Security Requirements Matrix}
\begin{table*}[t]
\centering
\small
\begin{tabularx}{\textwidth}{|X|X|X|}
\hline
\textbf{Requirement} & \textbf{CLI Assessment} & \textbf{MCP Assessment} \\
\hline
Least Privilege & Does not satisfy - all-or-nothing shell access & Satisfies - per-tool granular permissions \\
Credential Isolation & Does not satisfy - env vars visible to agent & Satisfies - server-held, opaque to model \\
Input Validation & Does not satisfy - vulnerable to command injection & Satisfies - JSON schema enforcement \\
Output Sanitization & Does not satisfy - stdout may leak secrets & Satisfies - controlled return values \\
Audit (SOC2) & Weak - shell history only (unstructured) & Satisfies - structured event logging \\
Data Classification & Does not satisfy - no data awareness & Satisfies - data labeling support \\
Human-in-the-Loop & Not supported - immediate execution & Satisfies - approval flow integration \\
Rate Limiting & Not supported - risk of resource exhaustion & Satisfies - per-user rate limits \\
\hline
\end{tabularx}
\end{table*}

\subsection{Compliance Framework Mapping}
\begin{table*}[t]
\centering
\small
\begin{tabularx}{\textwidth}{|X|X|X|X|}
\hline
\textbf{Framework} & \textbf{Key Requirement} & \textbf{CLI} & \textbf{MCP} \\
\hline
SOC 2 Type II & Access control + audit evidence & Manual, significant gaps & Native support \\
HIPAA & Minimum necessary access to PHI & Cannot restrict adequately & Per-field redaction \\
GDPR & Access logging + right to erasure & Partially achievable & Automated DPA compliance \\
PCI-DSS & Network segmentation & Flat network exposure & API boundary enforced \\
ISO 27001 & Risk management + access review & Achievable but expensive & Built-in controls \\
FedRAMP & Continuous monitoring + boundary & No protocol boundary & Protocol = authorization boundary \\
\hline
\end{tabularx}
\end{table*}

\begin{figure}[htbp]
\centering
\includegraphics[width=\columnwidth]{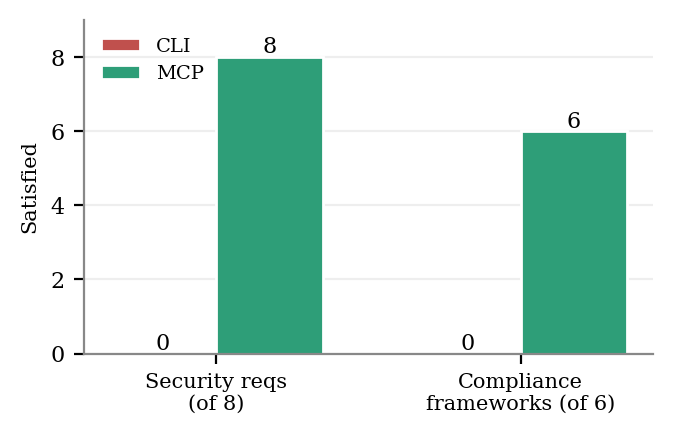}
\caption{Security and compliance scorecard: requirements satisfied (of 8) and frameworks met (of 6) by each paradigm.}
\end{figure}

\textit{Note:} The NSA published formal security guidance for MCP in May 2026 (CSI U/OO/6030316-26). No equivalent guidance exists for CLI-based AI agents because CLI provides no security architecture to govern. The absence is itself a risk indicator.

\section{Risk Analysis}
\subsection{CLI Architectural Risks}
\begin{table*}[t]
\centering
\small
\begin{tabularx}{\textwidth}{|X|X|X|X|}
\hline
\textbf{Risk} & \textbf{Attack Vector} & \textbf{Impact} & \textbf{Likelihood} \\
\hline
Shell Injection & Malicious command from prompt injection & Critical & Medium \\
Credential Theft & Agent reads \textasciitilde{}/.ssh/*, .env, env vars & Critical & High \\
Data Exfiltration & curl/wget to external endpoints & Critical & Medium \\
Destructive Commands & rm -rf, DROP TABLE, git push --force & High & Medium \\
Resource Exhaustion & Fork bomb, infinite loop, disk fill & High & Low-Med \\
Privilege Escalation & sudo, SUID, container escape & Critical & Low \\
SSRF & curl hits internal metadata endpoints & High & Medium \\
\hline
\end{tabularx}
\end{table*}

\subsection{MCP Operational Risks}
\begin{table*}[t]
\centering
\small
\begin{tabularx}{\textwidth}{|X|X|X|X|}
\hline
\textbf{Risk} & \textbf{Attack Vector} & \textbf{Impact} & \textbf{Mitigation} \\
\hline
Tool Poisoning & Malicious MCP server returns harmful instructions & High & Server allowlisting, signing \\
Schema Exploitation & Manipulated descriptions trick model & Medium & Trusted registries \\
Excessive Permissions & Overly broad tool scopes & Medium & Least-privilege RBAC \\
Token Overhead & Too many tools loaded & Low & Dynamic loading, filtering \\
\hline
\end{tabularx}
\end{table*}

\subsection{Underexplored Risks (New Analysis)}
\begin{table*}[t]
\centering
\small
\begin{tabularx}{\textwidth}{|X|X|X|X|}
\hline
\textbf{Risk Factor} & \textbf{CLI Impact} & \textbf{MCP Impact} & \textbf{Assessment} \\
\hline
State Leakage Between Calls & Environment variables, working directory, background processes persist across... & Each tool call is stateless by protocol design. No cross-call state leakage p... & MCPwins - architectural isolation \\
Tool Version Brittleness & CLI output format changes silently across OS versions (e.g.,lson macOS vs Lin... & Tool schemas are versioned. Breaking changes require explicit schema version ... & MCPwins - versioned contracts \\
Retry / Idempotency & No built-in retry semantics. Agent must implement retry logic manually. Non-i... & Server can declare idempotency keys. Protocol supports request IDs for dedupl... & MCPwins - protocol-level support \\
\hline
\end{tabularx}
\end{table*}

\textit{Key Distinction:} CLI risks are \textit{architectural} and cannot be fixed without abandoning CLI. MCP risks are \textit{operational} and have documented mitigations from NSA, OWASP, CoSAI, and CSA.

\section{Cost \& Token Economics}
\subsection{Per-Operation Token Cost}

    Early critiques of MCP focused on naive schema loading, where an entire server's tool manifest is injected into the context window upfront. Table V compares this naive case against progressive disclosure (Section IX) and the often-overlooked overhead of CLI-based agents.

\begin{table*}[t]
\centering
\small
\begin{tabularx}{\textwidth}{|X|X|X|X|}
\hline
\textbf{Cost Component} & \textbf{CLI Agent} & \textbf{MCP (Naive Loading)} & \textbf{MCP (Progressive Disclosure)} \\
\hline
Upfront schema overhead & 0 tokens\textsuperscript{\dag} & 2,000--55,000 tokens & \textasciitilde{}170 tokens (2 meta-tools) \\
Per-tool invocation & 50--200 tokens (command + flags) & 80--150 tokens (structured call) & 80--250 tokens (lookup + call) \\
Output parsing & 100--500 tokens (unstructured text) & 50--150 tokens (structured JSON) & 50--150 tokens (structured JSON) \\
System prompt / tool instructions & 200--600 tokens & Included in schema & Included in meta-tool schema \\
Error recovery (typical) & 300--1,200 tokens/retry & 100--300 tokens/retry & 100--300 tokens/retry \\
\hline
\end{tabularx}
\end{table*}

\textsuperscript{\dag} CLI agents carry no formal schema, but still require system-prompt instructions describing available commands, output formats, and error handling conventions. This overhead is frequently omitted from CLI benchmarks.

\subsection{Total Cost of Ownership (TCO)}
\begin{table*}[t]
\centering
\small
\begin{tabularx}{\textwidth}{|X|X|X|}
\hline
\textbf{Cost Factor} & \textbf{CLI} & \textbf{MCP} \\
\hline
Token cost / query & Lower (no schema) & Higher without progressive disclosure; comparable with it \\
Security incident cost & \$4.88M avg breach (IBM 2024) & Reduced attack surface \\
Compliance audit cost & Manual evidence (\$\$\$) & Automated exports \\
Multi-user infrastructure & Container/user (\$\$\$) & Single server + RBAC (\$) \\
Integration maintenance & Custom scripts per tool & Substantial reduction ($\sim$70\% claimed, BCG) \\
\hline
\end{tabularx}
\end{table*}

\begin{figure}[htbp]
\centering
\includegraphics[width=\columnwidth]{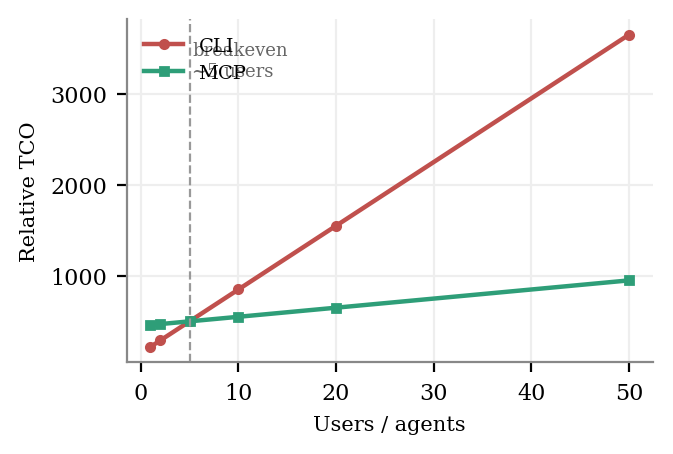}
\caption{Total cost of ownership vs.\ organization size. CLI is cheaper for small teams; MCP breaks even near five users and dominates at scale.}
\end{figure}

\subsection{Hidden Cost of Autonomous CLI Agents}

    The CLI token-efficiency argument assumes a cooperative, developer-present workflow. Autonomous CLI agents operating without human oversight face compounding costs that are frequently omitted from benchmarks:

\begin{table*}[t]
\centering
\small
\begin{tabularx}{\textwidth}{|X|X|X|}
\hline
\textbf{Failure Mode} & \textbf{Token Cost} & \textbf{Explanation} \\
\hline
Script generation & 200--800 tokens & Agent must compose multi-line shell scripts on the fly \\
Output parsing ambiguity & +150--400 tokens/retry & Unstructured stdout requires LLM interpretation; edge cases trigger retries \\
Error recovery loops & +300--1,200 tokens/attempt & Non-zero exit codes $\rightarrow$ agent re-plans, retries with different flags \\
Environment drift & +500--2,000 tokens & Commands fail due to missing tools, different OS, permission changes \\
Multi-step orchestration & +1,000--5,000 tokens & Complex workflows require piping, temp files, cleanup scripts \\
\hline
\end{tabularx}
\end{table*}

    In practice, autonomous CLI agents can consume an estimated \textbf{3--8$\times$ more tokens than idealized projections} due to retry loops and environment-specific failures; production agentic systems operating at hyperscale exhibit precisely these dynamics [24]. MCP's structured responses eliminate parsing ambiguity entirely, and typed error codes enable deterministic fallback without re-prompting the LLM [5].

\textit{Economics:} CLI has lower marginal token cost \textit{in ideal conditions}. Under autonomous operation with retries and error handling, MCP often achieves lower actual token spend. MCP has lower total cost of ownership at scale. Breakeven: >5 users or regulated data.

\section{Progressive Disclosure: Eliminating MCP's Primary Weakness}

    The CLI camp's strongest argument, that MCP floods the context window with thousands of tokens of tool schemas, is addressed by \textbf{progressive disclosure} [18]. This pattern exposes only a minimal tool registry at conversation start, loading full schemas on demand.

\subsection{The Problem}

    A typical MCP server (e.g., GitHub) ships 80 tools. At \textasciitilde{}700 tokens per schema, this costs \textbf{\textasciitilde{}55,000 tokens} injected before the agent performs any useful work. This is the core inefficiency that motivates the "CLI is enough" position.

\subsection{The Solution: Two Meta-Tools}

    Progressive disclosure replaces N tool schemas with exactly two meta-tools:

\begin{table*}[t]
\centering
\small
\begin{tabularx}{\textwidth}{|X|X|X|}
\hline
\textbf{Meta-Tool} & \textbf{Schema Cost} & \textbf{Function} \\
\hline
get\_tool(name) & \textasciitilde{}80 tokens & Returns the full schema for a named tool on demand \\
invoke\_tool(name, args) & \textasciitilde{}90 tokens & Executes a named tool with provided arguments \\
\hline
\end{tabularx}
\end{table*}

    Total upfront cost: \textbf{\textasciitilde{}170 tokens} (vs. 55,000 for full disclosure). The agent discovers tools as needed, matching CLI's token profile while retaining MCP's structured guarantees.

\subsection{Implementation: Solo.io Agent Gateway}

    Solo.io's \textit{agentgateway} implements progressive disclosure via a \texttt{toolMode: Search} configuration [18]:

\begin{lstlisting}
listeners:
  - name: mcp-listener
    targetGroups:
      - name: github-tools
        toolMode: Search    # <- progressive disclosure enabled
        tools:
          - name: "create_issue"
          - name: "list_pull_requests"
          - name: "get_file_contents"
          # ... 77 more tools available but NOT loaded into context
\end{lstlisting}

    When \texttt{toolMode: Search} is set, the gateway advertises only two tools to the LLM client. The agent uses \texttt{get\_tool} to discover specific tools when needed, then \texttt{invoke\_tool} to call them - paying schema cost only for tools actually used.

\subsection{Industry Benchmarks: Token Reduction}

    Multiple independent implementations have validated progressive disclosure in production, converging on 85-160x token reductions:

\begin{table*}[t]
\centering
\small
\begin{tabularx}{\textwidth}{|X|X|X|X|}
\hline
\textbf{Implementation} & \textbf{Technique} & \textbf{Token Reduction} & \textbf{Source} \\
\hline
Speakeasy Dynamic Toolsets & Semantic search +describe\_tools & 100--160$\times$(96\% avg.) & [19] \\
SynapticLabs Meta-Tool & Discovery + Execution (2 meta-tools) & 85--99\% & [20] \\
Kruczek Benchmark & On-demand schema fetch & 85$\times$ & [21] \\
Code Execution MCP (Brown) & Sandboxed Python replacing schemas & 98.7\% & [22] \\
Glama Token Elimination & Code execution replaces tool registry & 95--99\% & [23] \\
Solo.io AgentGateway & toolMode: Search & 71--97\% & [18] \\
\hline
\end{tabularx}
\end{table*}

\subsection{Token Impact: Worked Example}
\begin{table*}[t]
\centering
\small
\begin{tabularx}{\textwidth}{|X|X|X|X|}
\hline
\textbf{Scenario} & \textbf{Full Disclosure} & \textbf{Progressive} & \textbf{Savings} \\
\hline
GitHub server (80 tools), use 2 & 55,000 tokens & 170 + 1,400 = 1,570 & 97\% \\
File system (13 tools), use 2 & 2,100 tokens & 170 + 350 = 520 & 75\% \\
Database server (8 tools), use 3 & 4,200 tokens & 170 + 1,050 = 1,220 & 71\% \\
Multi-server (150 tools), use 5 & 105,000 tokens & 170 + 3,500 = 3,670 & 96\% \\
Enterprise (500+ tools), use 8 & 350,000+ tokens & 170 + 5,600 = 5,770 & 98.4\% \\
\hline
\end{tabularx}
\end{table*}

\subsection{Architectural Implications}

    Progressive disclosure transforms the CLI-vs-MCP cost comparison. The SynapticLabs "three-layer architecture" [20] organizes this into: (1) Meta-Tools as entry points (2 tools registered), (2) Bounded Context Packs grouping tools by domain (following the 7±2 cognitive limit), and (3) Individual tool schemas loaded only on invocation. With this pattern active:

\begin{itemize}
  \item \textbf{Token efficiency:} MCP matches or beats CLI (170 tokens upfront vs. CLI's zero - negligible difference)
  \item \textbf{Security:} MCP retains full RBAC, audit, and credential isolation
  \item \textbf{Scalability:} Adding 100 new tools costs zero additional context tokens
  \item \textbf{Discovery:} Agent can ask "what tools are available for GitHub?" without pre-loading all schemas
\end{itemize}

\textit{Progressive Disclosure:} Eliminates MCP's primary cost disadvantage (context window bloat) while retaining all security and governance benefits. With this pattern, MCP achieves CLI-equivalent token efficiency for the first time.

\section{Industry Adoption \& Evidence}
\begin{figure}[htbp]
\centering
\includegraphics[width=\columnwidth]{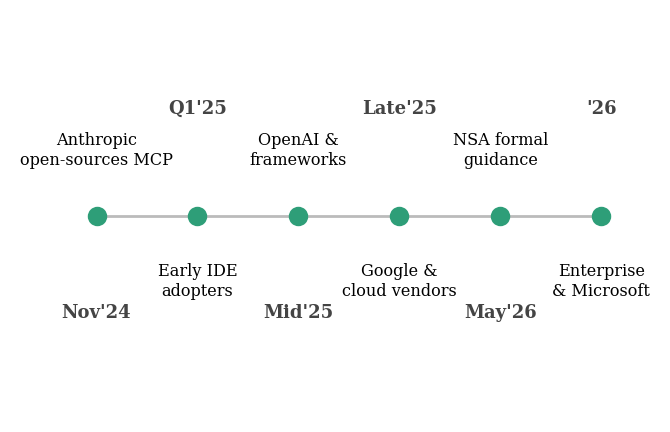}
\caption{MCP adoption timeline: major vendors adopted the protocol within roughly twelve months of its late-2024 open-source release.}
\end{figure}

\begin{table*}[t]
\centering
\small
\begin{tabularx}{\textwidth}{|X|X|X|}
\hline
\textbf{Organization} & \textbf{MCP Role} & \textbf{Year} \\
\hline
Anthropic & Created MCP; open-sourced specification & 2024 \\
OpenAI & Adopted in ChatGPT \& Agent SDK & 2025 \\
Google DeepMind & MCP support in Gemini/Vertex & 2025 \\
Microsoft & MCP in Copilot, Azure AI Foundry & 2025 \\
AWS & MCP in Bedrock agent framework & 2025 \\
Linux Foundation & MCP donated to Agentic AI Foundation & 2025 \\
NSA & Published formal security guidance & 2026 \\
\hline
\end{tabularx}
\end{table*}

\begin{table*}[t]
\centering
\small
\begin{tabularx}{\textwidth}{|X|X|X|}
\hline
\textbf{Metric} & \textbf{Value} & \textbf{Source} \\
\hline
Enterprise apps w/ task-specific AI agents (2026) & 40\% & Gartner [14] \\
Integration cost reduction & up to $\sim$70\% & BCG [12] \\
Enterprise apps w/ such agents (2025 baseline) & \textless{}5\% & Gartner [14] \\
MCP ecosystem connectors & 5,800+ & MCP Registry \\
\hline
\end{tabularx}
\end{table*}

\section{Decision Framework \& Recommendation}
\subsection{Decision Matrix}
\begin{table*}[t]
\centering
\small
\begin{tabularx}{\textwidth}{|X|X|X|}
\hline
\textbf{Scenario} & \textbf{Recommendation} & \textbf{Rationale} \\
\hline
Solo developer, local, personal projects & CLI & Fastest, cheapest, no governance needed \\
Prototyping / hackathon & CLI & Speed over governance \\
CI/CD pipeline (trusted) & CLI+MCP & CLI for known commands; MCP for external APIs \\
Team of 5+ sharing agents & MCP & Per-user permissions mandatory \\
Production data access & MCP & Credential isolation required \\
Regulated industry & MCP & Compliance controls mandatory \\
Customer-facing AI product & MCP & Multi-tenant isolation, rate limiting \\
Enterprise (50+ users, SOC2) & MCP & No viable alternative at scale \\
\hline
\end{tabularx}
\end{table*}

\subsection{The Hybrid Pattern (Recommended)}
\begin{quote}
\small
Use CLI For:Local file system operationsBuild commands (npm, make, cargo)Git on local reposText processing (grep, sed, awk)Developer-facing operationsUse MCP For:External API access (GitHub, Jira)Database queries (production)Cloud resource managementActions requiring audit/approvalMulti-user / multi-tenantCredential-bearing operations
\end{quote}
\begin{figure}[htbp]
\centering
\includegraphics[width=\columnwidth]{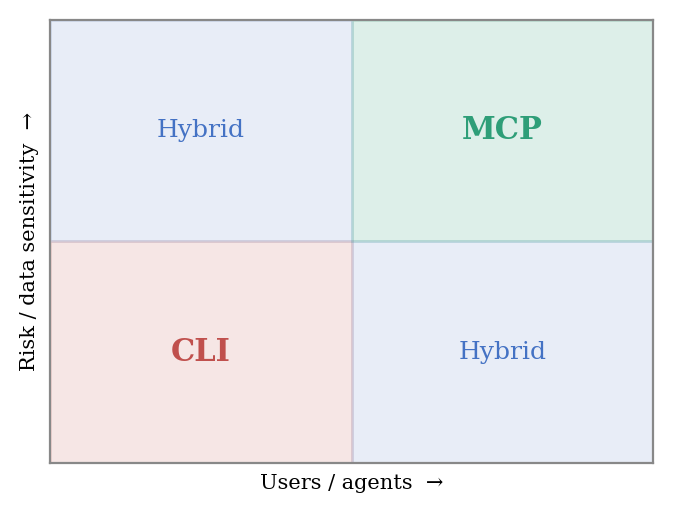}
\caption{Risk/scale matrix: CLI suits small, low-risk local use; MCP suits large, high-sensitivity multi-tenant use; a hybrid covers the middle.}
\end{figure}

\section{Conclusion}

    This analysis establishes that the CLI-vs-MCP debate is a \textbf{false dichotomy}. The two paradigms serve fundamentally different trust levels within an AI agent architecture:

\begin{itemize}
  \item \textbf{CLI excels} in low-risk, single-user, local operations where speed and simplicity are paramount (3/14 dimensions).
  \item \textbf{MCP excels} in every dimension relevant to enterprise deployment - security, compliance, audit, multi-tenancy, composability, and maintainability (9/14 dimensions). Progressive disclosure [18]--[23] eliminates the context window cost objection, making token efficiency a tie rather than a CLI advantage.
\end{itemize}

    The recommended architecture is \textbf{hybrid}: CLI for local developer workflows within the trust boundary; MCP for anything crossing trust boundaries, touching production data, or requiring governance. Organizations that use "CLI is enough" as justification to skip MCP are trading short-term token savings (\textasciitilde{}\$0.002/query) for long-term security debt (avg. breach cost: \$4.88M).

\textit{Final Recommendation:} Adopt MCP for all governed, multi-user, or production-facing AI agent interactions. Retain CLI for local developer tooling. This hybrid pattern captures CLI's efficiency without sacrificing MCP's security architecture.

\end{document}